\documentclass[runningheads]{llncs}
\usepackage[T1]{fontenc}
\usepackage{pifont}

\usepackage{amsmath,mathtools,amssymb,amsfonts,bm}

\usepackage{subcaption}
\usepackage{graphicx}
\usepackage{caption}
\usepackage{booktabs}
\usepackage{multirow}
\usepackage{hyperref}
\hypersetup{
    colorlinks=true,
    linkcolor=blue,
    filecolor=cyan,      
    urlcolor=magenta,
    }
\usepackage[nameinlink]{cleveref}
\usepackage{color}
\usepackage[table]{xcolor}

\newcommand{\rulesep}{\unskip\hfill{\color{black}\vrule width 1pt}\hfill\ignorespaces}

\begin{document}

\title{Unlocking Fine-Grained Details with Wavelet-based High-Frequency Enhancement in Transformers}
\titlerunning{Frequency Enhanced Transformer (FET)}

\author{Reza Azad\inst{1}\and
Amirhossein Kazerouni\inst{2}\and
Alaa Sulaiman\inst{3} \and
Afshin Bozorgpour\inst{4} \and
Ehsan {Khodapanah Aghdam} \inst{5}\and
Abin Jose \inst{1}\and
Dorit Merhof\inst{4}}

\institute{Faculty of Electrical Engineering and Information Technology, RWTH Aachen University, Germany \and
School of Electrical Engineering, Iran University of Science and Technology, Iran \and
Faculty of Information Science and Technology, Universiti Kebangsaan, Malaysia \and
Faculty of Informatics and Data Science, University of Regensburg, Germany \and
Department of Electrical Engineering, Shahid Beheshti University, Iran
\\\email{dorit.merhof@informatik.uni-regensburg.de}} 

\authorrunning{Azad et al.}% Part of LEFT running header
\maketitle              % typeset the header of the contribution

\newcommand\ea[1]{{\color{red}#1}}

\begin{abstract}
Medical image segmentation is a critical task that plays a vital role in diagnosis, treatment planning, and disease monitoring. Accurate segmentation of anatomical structures and abnormalities from medical images can aid in the early detection and treatment of various diseases.
In this paper, we address the local feature deficiency of the Transformer model by carefully re-designing the self-attention map to produce accurate dense prediction in medical images. To this end, we first apply the wavelet transformation to decompose the input feature map into low-frequency (LF) and high-frequency (HF) subbands. The LF segment is associated with coarse-grained features, while the HF components preserve fine-grained features such as texture and edge information. Next, we reformulate the self-attention operation using the efficient Transformer to perform both spatial and context attention on top of the frequency representation. Furthermore, to intensify the importance of the boundary information, we impose an additional attention map by creating a Gaussian pyramid on top of the HF components. Moreover, we propose a multi-scale context enhancement block within skip connections to adaptively model inter-scale dependencies to overcome the semantic gap among stages of the encoder and decoder modules. Throughout comprehensive experiments, we demonstrate the effectiveness of our strategy on multi-organ and skin lesion segmentation benchmarks. The implementation code will be available upon acceptance. \href{https://github.com/mindflow-institue/WaveFormer}{GitHub}.

\end{abstract}

\section{Introduction}\label{sec:intro}

In the field of computer vision, Convolutional Neural Networks (CNNs) have been the dominant architecture for various tasks for many years \cite{karimijafarbigloo2023self,molaei2023implicit}. More recently, however, the Vision Transformer (ViT)~\cite{dosovitskiy2020image} has been shown to achieve state-of-the-art (SOTA) results in diverse tasks with significantly fewer parameters than traditional CNN-based approaches. This has resulted in a shift in the field towards utilizing ViT, which is becoming increasingly popular for a wide range of computer vision tasks \cite{azad2023advances,bozorgpour2023dermosegdiff}. The main success behind the ViTs is their ability to model long-range contextual dependencies by applying a grid-based self-affinities calculation on image patches (tokens). Unlike CNNs, which require stacked convolution blocks to increase the receptive field size, the ViT captures the global contextual representation within a single block. However, the ViT model usually suffers from a weak local description compared to the CNN models, which is crucial for semantic segmentation tasks in medical images.

To address the local feature deficiency of Transformer models, recent studies have explored the combination of CNN-Transformer models or pure Transformer-based designs with U-Net-like architectures~\cite{chang2021transclaw,huang2022missformer}.
The strength of the U-Net lies in its symmetrical hierarchical design with a large number of feature channels. However, a pure Transformer-based design involves quadratic computational complexity of the self-attention operation with respect to the number of patches, which makes a combination of U-Net and Transformer challenging. Furthermore, due to this fixed-size scale paradigm, ViT has no strong spatial inductive bias. Therefore, extensive research endeavors aim to overcome these issues by designing efficient and linear complexity self-attention mechanisms to make ViTs suitable for dense prediction tasks. Such designs either diminish the patch numbers (e.g., ATS~\cite{renggli2022learning} or A-ViT~\cite{yin2022vit}), or apply downsampling or pooling operations, i.e., on images or key/value tensors (e.g., SegFormer~\cite{xie2021segformer}, PVT~\cite{wang2022pvt}, or MViT~\cite{fan2021multiscale}). Furthermore, calculation on self-attentions is hindered by local windowing schemas as in studies such as Swin Transformer~\cite{liu2021swin} or  DW-ViT~\cite{ren2022beyond}.
% Many attempts to take advantage of ViTs have been made as the main backbone counterparts in U-shaped pipelines for medical image segmentation. 
Swin-Unet~\cite{cao2021swinunet} explored the linear Swin Transformer in a U-shaped structure as a Transformer-based backbone. MISSFormer~\cite{huang2022missformer} investigated the efficient self-attention from SegFormer as a main module for 2D medical image segmentation. In contrast, these methods endorse the ability of ViTs in segmentation tasks but still suffer from boundary mismatching and poor boundary localization due to the information dropping through their enhanced and efficient self-attention process. On the other hand, the Swin Transformer~\cite{liu2021swin} utilizes non-overlapping windows to employ the self-attention mechanism which, however, may lead to the loss of detailed edges and other spatial information. Efficient self-attention~\cite{xie2021segformer} used in~\cite{huang2022missformer} decreases the dimensions of the input sequence in spatial dimensions that lose informative details and make the segmentation results error-prone. Moreover, recent studies~\cite{wang2022antioversmoothing} investigated how self-attention performs as a low-pass filter when Transformer blocks are stacked successively. Therefore, stacking Transformer blocks in a multi-scale paradigm (e.g. U-Net architecture) not only helps to model a multi-scale representation but also degrades the loss of local texture and localization features (high-frequency details) through the network.

High-frequency components are often critical in many real-world signals, such as speech and images, and they are usually associated with fine-grained details that can provide valuable information for many vision-based tasks. However, the Transformer model is known to consider low-frequency representations, making it challenging to capture these high-frequency components~\cite{wang2022antioversmoothing}. This limitation can result in vague and unsatisfactory feature extraction, leading to a suboptimal performance on the segmentation tasks, which requires a precise boundary extraction. Therefore, exerting wavelet analysis to enhance high-frequency representations in a Transformer can provide a multi-resolution decomposition of the input data, allowing us to identify and isolate high-frequency components that provide a more comprehensive representation.

In this paper, we propose a new Wavelet-based approach for medical image segmentation in a U-shaped structure with the help of efficient Transformers that modifies the quadratic self-attention map calculation by reformulating the self-attention map into a linear operation. We also propose incorporating a boundary attention map to highlight the importance of edge information further to distinguish overlapped objects, termed \textbf{F}requency \textbf{E}nhancement \textbf{T}ransformer (\textbf{FET}) block. Furthermore, we design an MSCE module within the skip connections to overcome the semantic gap among the encoder and decoder stages to build rich texture information transferring, which is otherwise limited by the multi-scale representation in a conventional encoder-to-decoder path. 
Our contributions are as \ding{182} We propose a novel FET block comprising a frequency-enhanced module and boundary-aware attention map to model both shape and texture representation in an adaptive way. \ding{183} Applying our proposed MSCE module to skip connections induces the informative texture information from the encoder to the decoder to enrich the missing localization information regarded as a low-frequency representation. \ding{184} In addition, our method leverages the high-frequency components after applying a Gaussian kernel to perform additional attention information that could effectively highlight the boundary and detailed information for dense prediction tasks, \textit{e.g.} segmentation.

\section{Proposed Method}\label{sec:method}

As illustrated in \Cref{fig:proposed-method}, our proposed method trains in an end-to-end strategy that incorporates the frequency analysis in a multi-scale representation within the efficient Transformer paradigm. Therefore, this section first recapitulates the seminal vision Transformer's inner structure by investigating the multi-head self-attention (MHSA) general mathematical formulation. Assume $X \in \mathbb{R}^{H\times W \times D}$ to be the 2D input image (or feature map stream), then $X$ can be reshaped as a sequence of patches consisting of $n = H \times W$ image patches, where $D$ is the dimension of each patch. Afterward, three representations are learned from the $X$, namely $Q \in \mathbb{R}^{n \times D}$ Queries, $K \in \mathbb{R}^{n \times D}$ Keys, and $V \in \mathbb{R}^{n \times D}$ Values. The multi-head attention regime utilizes $N_h$ diverse Queries, Keys, and Values, where $\left\{  Q_j,  V_j,  K_j  \right\} \in \mathbb{R}^{n \times {D_h}} $ depicts the $j$-th head information. Then, the MHSA follows and learns the final attention over calculated queries, keys, and values according to the following equations:
\begin{figure}[!th]
    \centering
    \includegraphics[width=\textwidth]{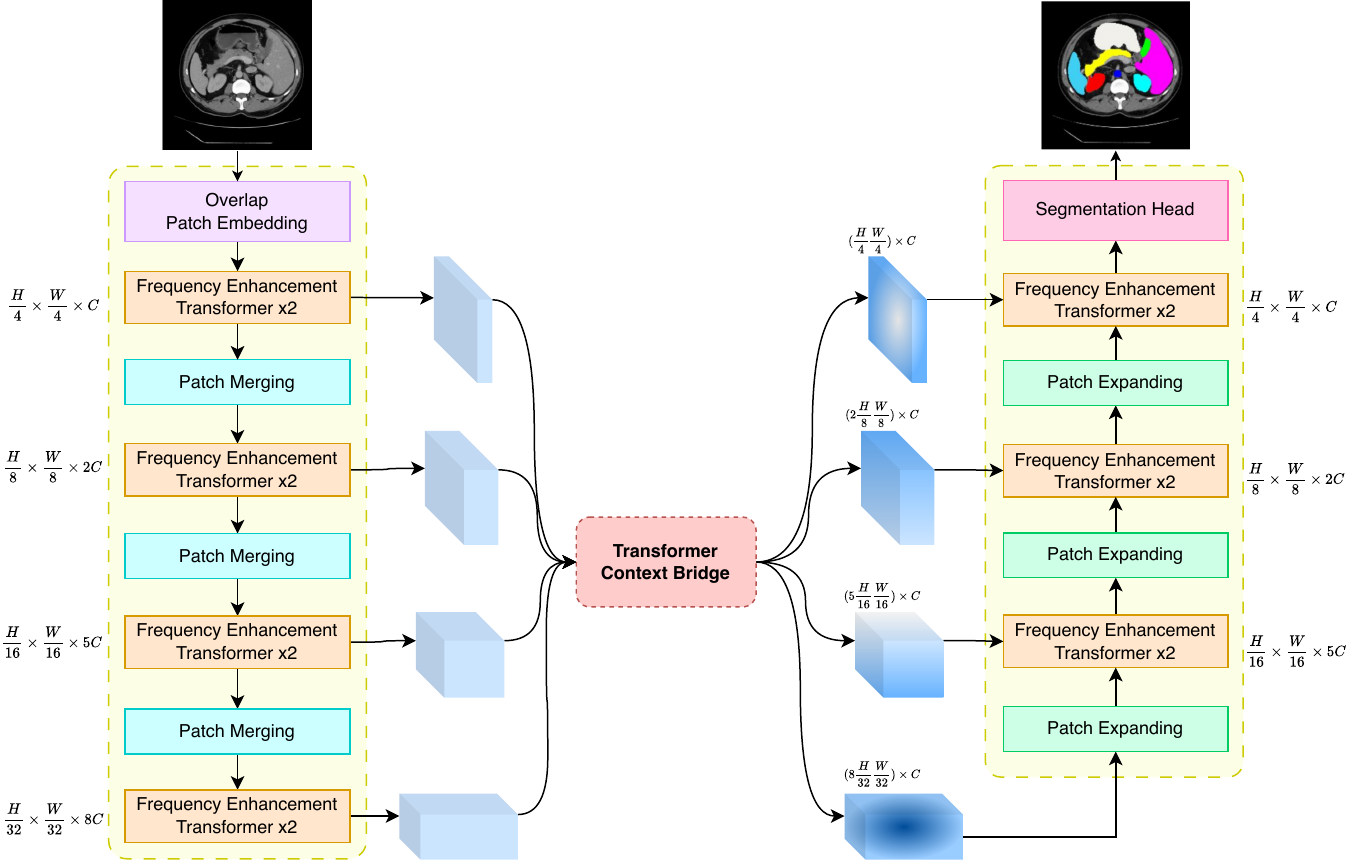}
    \caption{The overview of the proposed \textbf{F}requency \textbf{E}nhanced \textbf{T}ransformer (\textbf{FET}) model. Each frequency-enhanced Transformer block comprises the sequential LayerNorm, FET block, LayerNorm, and Mix-FFN modules.}
    \label{fig:proposed-method}
\end{figure} 
{\small
\begin{align}
\label{eq:MHSA}
\text{\textbf{MHSA}}(Q,K,V) &= \text{\textbf{Concat}}(head_0, head_1,...,head_{N_h})W^O, \notag\\
head_j &= \text{\textbf{Attention}}(Q_j,K_j,V_j), \notag \\
\text{\textbf{Attention}}(Q_j,K_j,V_j) &= \text{\textbf{Softmax}}(\frac{Q_j K_j^T}{\sqrt D_h})V_j,
\end{align}
}%
where \textbf{Concat} and $W^O$ denote the concatenation operation and the learnable transformation tensor, respectively. Thus, the conventional Transformer captures long-range dependencies but still suffers from several limitations that could affect the ViT's performance in dense segmentation tasks: first, the computational cost of multi-head self-attention is quadratic in patch numbers, $\mathcal{O}(n^2D)$, making it unsuitable for high-resolution tasks. Second, the recent analytic work from Wang et al.~\cite{wang2022antioversmoothing} demonstrated the deficiency of a multi-head self-attention mechanism in capturing high-frequency details due to the included \textbf{Softmax} operation. Specifically, the lack of ability to capture high-frequency information degrades the segmentation performance with naive ViTs. Therefore, in the next section, we propose our FET module to address all aforementioned issues.

\subsection{Efficient Transformer}

Due to the quadratic computational complexity of seminal Transformers, a wide range of studies have been conducted to minimize this weakness. Shen et al.~\cite{shen2021efficient} revisited the dot production within the multi-head self-attention mechanism to circumvent redundant operations. From \Cref{eq:MHSA}, it can be seen that the MHSA captures the similarity between each pair of patches, which is much more resource intensive. Efficient attention computes the self-attention as
\begin{align}
    \text{\textbf{Efficient Attention}} = \bm{\rho_q}(Q)(\bm{\rho_k} (K)^TV), \label{eq:eff_MHSA}
\end{align}
where $\bm{\rho_q}$ and $\bm{\rho_k}$ denote the normalization functions for $Q$ and $K$. In \Cref{eq:eff_MHSA}, instead of considering the keys as $n$ feature vectors in $\mathbb{R}^D$, the module interprets them as $d_k$ feature maps with only one channel. Efficient attention applies these feature maps as weights across all positions and combines the value features by weighted summation, resulting in a global context vector. This vector does not refer to any particular position, but rather represents a comprehensive overview of the input features, analogous to a global context vector.

\subsection{Frequency Enhancement Transformer (FET)}

As suggested by~\cite{yao2022wave}, we follow their intuition to preserve the high-frequency counterparts for medical image segmentation tasks. Discrete Wavelet Transform (DWT) is a mapping function from spatial resolution to spatial-frequency space. Wavelet decomposition is a powerful technique that decomposes images into high and low-frequency components, providing a multi-resolution analysis of the input signal. In medical image segmentation, high-frequency components of the image correspond to fine details such as edges and texture. In contrast, low-frequency components correspond to large-scale structures and background information. Thus, a wavelet decomposition which analyzes both high and low-frequency components of medical images may enhance the accuracy of segmentation models by capturing both local and global features of the image. While applying DWT on an image, there would be four distinct wavelet subbands, namely LL, LH, HL, and HH, demonstrating the texture, horizontal details, vertical details, and diagonal information, respectively. 
% Each subband contains information about the image's different frequency ranges and spatial resolutions. 

\begin{figure*}[!th]
\centering
\begin{subfigure}{0.48\textwidth}
    \centering
    \includegraphics[width=0.99\textwidth]{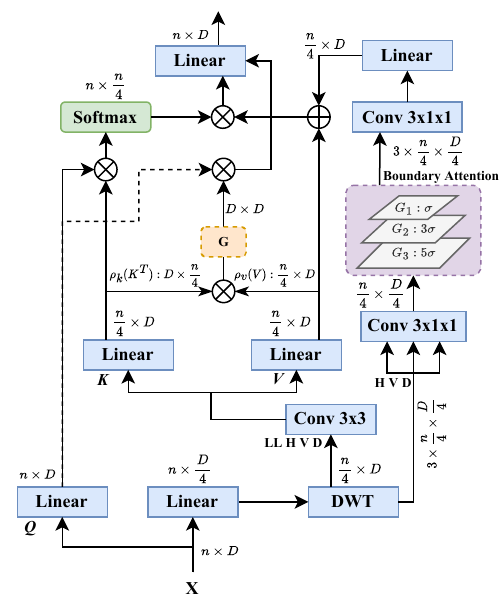}
    \caption{}
    \label{fig:proposed_module}
\end{subfigure}
\rulesep
\begin{subfigure}{0.48\textwidth}
    \centering
    \includegraphics[width=0.95\textwidth]{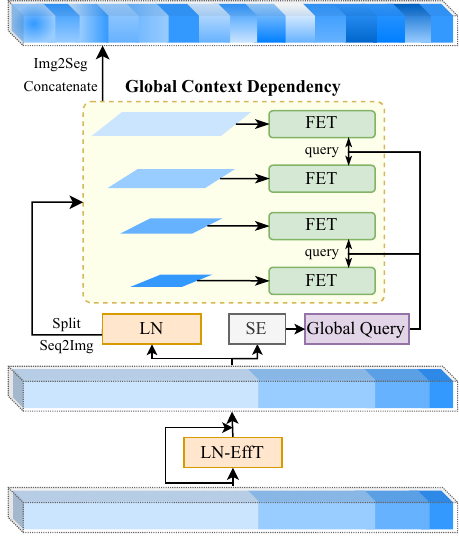}
    \caption{}
    \label{fig:proposed_skipconnection_module}
\end{subfigure}
\vspace{-0.75em}
\caption{\textbf{(a)} The \textbf{FET} Block. [LL, H, V, D] denotes the low-frequency, horizontal, vertical, and diagonal high-frequency counterparts. \textbf{(b)} The overview of \textbf{MSCE} skip connection enhancement module. LN, EffT, and SE are the LayerNorm, the efficient Transformer module, and the squeeze and excitation block, respectively.}

\label{fig:mainfig}
\end{figure*}

The FET (visualized in \Cref{fig:proposed_module}) is designed to address previous limitations by highlighting the boundary information (high-frequency details) for medical image segmentation. Motivated by~\cite{yao2022wave}, FET utilizes the DWT to account for the frequency analysis for focusing on high-frequency counterparts. First, the input 2D image (feature map) $X \in \mathbb{R}^{H \times W \times D}$ ($n = H \times W$) is linearly transformed into $\overset{\sim}{X} = \mathbb{R}^{n \times \frac{D}{4}}$ by reducing the channel dimension. Classical DWT applies pairs of low-pass and high-pass filters along rows and columns to extract frequency response subbands. Next, DWT is applied to $\overset{\sim}{X}$ to extract frequency responses and to downsample the input. As a result, the four subbands of input are $\overset{\sim}{X} = [\overset{\sim}{X}_{LL}, \overset{\sim}{X}_{LH}, \overset{\sim}{X}_{HL}, \overset{\sim}{X}_{HH}] \in \mathbb{R}^{n \times \frac{D}{4}}$. The high-frequency components ($\overset{\sim}{X}_{LH}$, $\overset{\sim}{X}_{HL}$, and $\overset{\sim}{X}_{HH}$) concatenate in a new dimension due to the underlying texture details at the fine-grained level. Then, a $3\times 1 \times 1$ convolution is applied to the resulting feature map to recalibrate for a subsequent Gaussian hierarchical ``Boundary Attention'' mechanism. The process continues with another $3 \times 1 \times 1$ convolution, and then the encoded boundary features are concatenated in the channel dimension. Analogous to~\cite{yao2022wave}, another branch applies a $3 \times 3$ convolution for creating the keys and values. Furthermore, a global context results from incorporating keys and values. However, to compensate for the \textbf{Softmax} operation's destructive effect~\cite{wang2022antioversmoothing}, we add the boundary attention to Value, to include the boundary preservation action when calculating attention. After boundary extraction, the FET block uses a query $Q$ from the input $X$ and key $K$ and value $V$ from the DWT to extract multi-disciplinary contextual correlations. While the firstmost left branch captures the spatial dependencies, the middle branch extracts the channel representation in an efficient concept. In addition, the most right branch highlights the boundary information within the value representation. Finally, the FET model in \Cref{fig:proposed-method} is composed of a LayerNorm, FET block (see \Cref{fig:proposed_module}), LayerNorm, and Mix-FFN~\cite{xie2021segformer} modules in sequence.

\subsection{Multi-Scale Context Enhancement (MSCE)}

A multi-scale fusion paradigm is considered in our design for accurate semantic segmentation to alleviate the semantic gap between stages of U-shaped structures, as in \Cref{fig:proposed_skipconnection_module}. Given the multi-level features that resulted from the hierarchical encoder, representations are flattened in spatial dimension and are reshaped to keep the same channel depth at each stage. Considering $F_i$ as a hierarchical feature in each encoder stage $i \in \{1,\dots,4\}$, we flatten them spatially and reshape them to obtain the same channel depth for each stage before concatenating them in the spatial dimension. Following the LayerNorm and efficient Transformer, we create the hierarchical long-range contextual correlation. Afterward, the tokens are split and reshaped to their original shape of features in each stage and are fed to the FET block to capture the amalgamated hierarchical contextual representation. We capture the global information from the represented token space as a \textit{Global Query} to the FET blocks.

\section{Experiments}\label{sec:experiments}

Our proposed method was implemented using the PyTorch library and executed on a single RTX 3090 GPU. A batch size of \textcolor{black}{24} and a SGD solver with a base learning rate of \textcolor{black}{0.05}, a momentum of \textcolor{black}{0.9}, and a weight decay of \textcolor{black}{0.0001} is used. The training was carried out for \textcolor{black}{400} epochs. For the segmentation task, both cross-entropy and Dice losses were utilized as the loss function. The segmentation task was performed using the combined loss ($Loss = 0.6 \cdot L_{dice} + 0.4 \cdot L_{ce}$ as used in \cite{heidari2023hiformer}).
\noindent\textbf{Datasets:}\label{sec:dataset}
First, we evaluated our method on the \textit{Synapse} dataset~\cite{landman2015miccai} that contains 30 cases of abdominal CT scans with 3,779 axial contrast-enhanced abdominal clinical CT images. Each CT data consists of $85 \sim 198$ slices of a consistent size $512 \times 512$  with the eight organ classes annotation. We followed the same preferences for data preparation as in~\cite{chen2021transunet}. Second, our study on skin lesion segmentation is based on the \textit{ISIC 2018}~\cite{codella2019skin} dataset, which was published by the International Skin Imaging Collaboration (ISIC) as a large-scale dataset of dermoscopy images. We follow the~\cite{azad2019bi} for the evaluation setting.

\noindent\textbf{Qualitative and Quantitative Results:}\label{sec:results}
In \Cref{comparison_synapse}, we compare the performance of our proposed FET method with previous SOTA methods for segmenting abdominal organs using the DSC and the HD metrics. Our method surpasses existing CNN-based methods by a significant margin. FET exhibits superior learning ability on the DSC metric compared to other models, achieving an increase of 1.9\% compared to HiFormer. The quantitative results highlight the FET superiority in segmenting kidney, pancreas, and spleen organs. The \Cref{fig:synapseviz} also endorses the mentioned results qualitatively, and all other models suffer from organ deformations when segmenting the liver and suffer from under-segmentation while FET performs smoothly.

\begin{table}[!thb]
    \centering
    \caption{Comparison results of the proposed method on the \textit{Synapse} dataset. \textcolor{blue}{Blue} indicates the best result, and \textcolor{red}{red} displays the~second-best.}
    \label{comparison_synapse}
    \resizebox{1\textwidth}{!}{
    \begin{tabular}{l|c|cc|cccccccc} 
    \toprule
    \textbf{Methods} & \textbf{\# Params (M)} & \textbf{DSC~$\uparrow$} & \textbf{HD~$\downarrow$} & \textbf{Aorta} & \textbf{Gallbladder} & \textbf{Kidney(L)} & \textbf{Kidney(R)} & \textbf{Liver} & \textbf{Pancreas} & \textbf{Spleen} & \textbf{Stomach} \\ 
    \midrule
    R50 U-Net~\cite{chen2021transunet} & 30.42 & 74.68 & 36.87 & 87.74 & 63.66 & 80.60 & 78.19 & 93.74 & 56.90 & 85.87 & 74.16 
    \\
    U-Net~\cite{ronneberger2015u} & 14.8 &76.85 & 39.70 & \textcolor{red}{89.07} & \textcolor{blue}{69.72} & 77.77 & 68.60 & 93.43 & 53.98 & 86.67 & 75.58 
    \\
    % R50 Att-UNet~\cite{chen2021transunet}   & 75.57 & 36.97 & 55.92 & 63.91 & 79.20 & 72.71 & 93.56 & 49.37 & 87.19 & 74.95 
    % \\
    Att-UNet~\cite{schlemper2019attention}  &  34.88 & 77.77 & 36.02 & \textcolor{blue}{89.55} & 68.88 & 77.98 & 71.11 & 93.57 & 58.04 & 87.30 & 75.75 
    \\
    % R50 ViT~\cite{chen2021transunet}  & 71.29 & 32.87 & 73.73 & 55.13 & 75.80 & 72.20 & 91.51 & 45.99 & 81.99 & 73.95 
    % \\
    TransUnet~\cite{chen2021transunet} &  105.28  & 77.48 & 31.69 & 87.23 & 63.13 & 81.87 & 77.02 & 94.08 & 55.86 & 85.08 & 75.62 
    \\
    Swin-Unet~\cite{cao2021swinunet} &  27.17  & 79.13 & 21.55 & 85.47 & 66.53 & 83.28 & 79.61 & 94.29 & 56.58 & 90.66 & 76.60 
    \\
    LeVit-Unet~\cite{xu2021levit} & 52.17  & 78.53 & \textcolor{red}{16.84} & 78.53 & 62.23 & 84.61 & \textcolor{blue}{80.25} & 93.11 & 59.07 & 88.86 & 72.76 
    \\
    DeepLabv3+ (CNN)~\cite{chen2018encoder} &  59.50 & 77.63 & 39.95 & 88.04 & 66.51 & 82.76 & 74.21 & 91.23 & 58.32 & 87.43 & 73.53 
    \\
    HiFormer~\cite{heidari2023hiformer}  & 25.51 & 80.39  &  \textcolor{blue}{14.70} & 86.21  & 65.69   & \textcolor{red}{85.23}  & 79.77   & \textcolor{blue}{94.61}  & 59.52   & \textcolor{red}{90.99}  & \textcolor{red}{81.08}
    \\
    \midrule
    Baseline &   27.36  &   80.39 & 20.56 & 85.69 & \textcolor{red}{69.68} & 83.83 & \textcolor{red}{80.07} & 94.20 & 60.72 & 90.92 & 77.98 
    \\
    FET (without MSCE bridge) &   33.00   & \textcolor{red}{81.05} & 17.70 & 87.80 & 68.33 & 85.00 & 79.25 & 94.11 & \textcolor{red}{61.80} & 90.95 & \textcolor{blue}{81.18}
    % \\
    % \rowcolor[HTML]{C8FFFD}
    % \textbf{FET} &  39.02  &\textcolor{blue}{82.01} & \textcolor{red}{16.34} & 88.65 & \textcolor{blue}{69.96} & \textcolor{red}{84.93} & \textcolor{red}{80.22} & \textcolor{blue}{94.37} & \textcolor{blue}{62.33} & \textcolor{blue}{91.55} & \textcolor{blue}{82.83} 
    \\
    \rowcolor[HTML]{C8FFFD}
    \textbf{FET} &  47.01  &\textcolor{blue}{81.92} & 18.41 & 85.31 & 69.67 & \textcolor{blue}{86.66} & 80.06 & \textcolor{red}{94.43} & \textcolor{blue}{67.08} & \textcolor{blue}{91.85} & 80.34
    \\
    \bottomrule
    \end{tabular}
    }
\end{table}

\begin{table}[!thb]
    \caption{(a) Segmentation results of the proposed method versus SOTA methods on the \textit{Synapse} dataset. (b) Quantitative results on ISIC2018 dataset.}
    
    % \vspace{-0.85em}
    \begin{subtable}[h]{0.65\textwidth}
        % \begin{figure}
            \centering
            \caption{Segmentation visualization on Synapse dataset.}
        	\vspace{-.8em}
            \includegraphics[width=\textwidth]{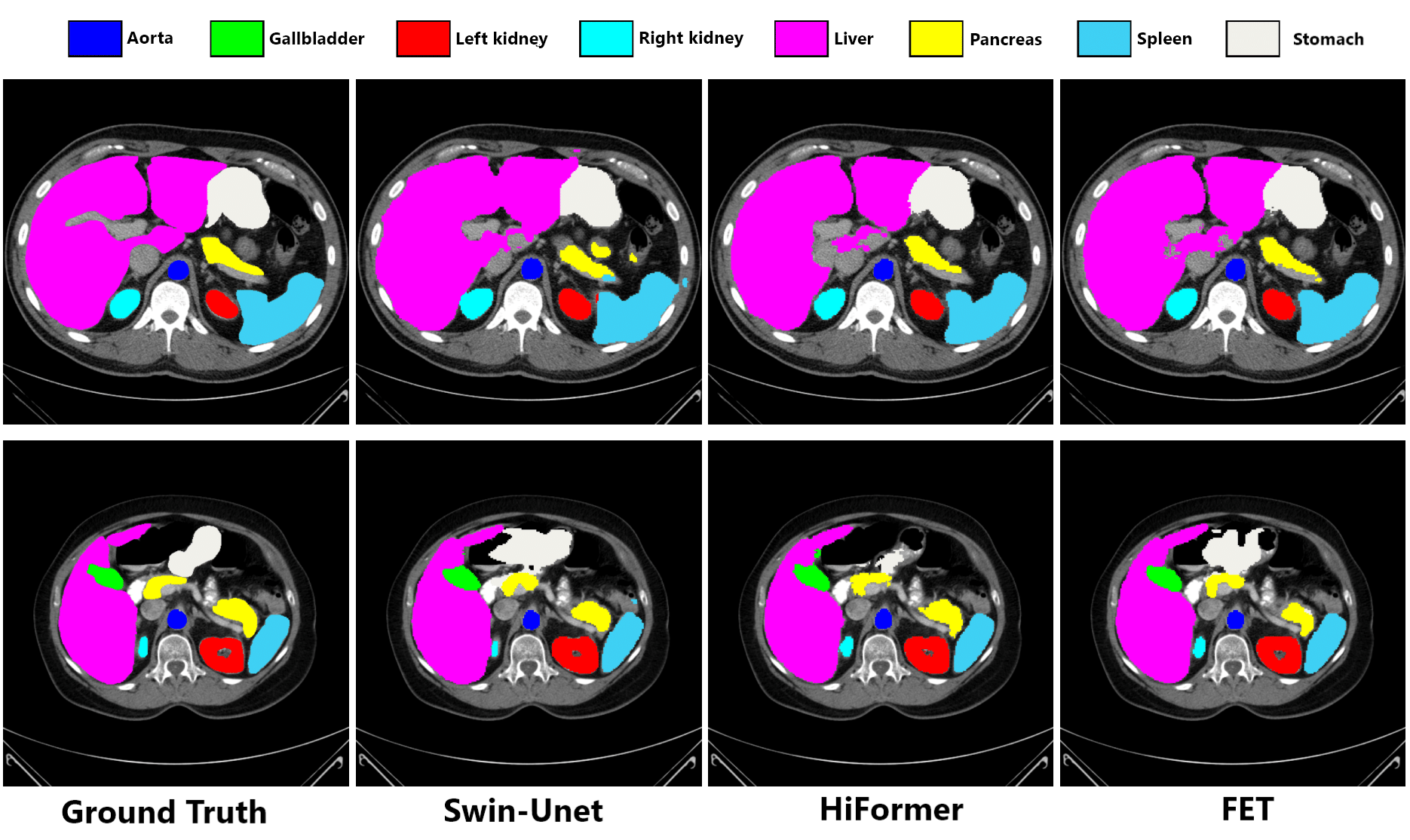}
            % {
            %     \tiny
            %     Ground Truth \hspace{2.4em} 
            %     Swin-UNet \hspace{2.8em} MISSFormer \hspace{3.4em}
            %     UNet
            %     \hspace{1.2em}
            % }
            \label{fig:synapseviz}
        % \end{figure}
    \end{subtable}
    \hfill
    \begin{subtable}[h]{0.34\textwidth}
    \centering
    \caption{\textit{ISIC 2018}}
    \label{tab:skin_comparison}
    \resizebox{\textwidth}{!}{
    \begin{tabular}{l||cccc} 
    \toprule
    \multirow{2}{*}{\textbf{Methods}}  &  \multicolumn{4}{c}{\textbf{ISIC 2018}} \\ 
    \cline{2-5}
     & \textbf{DSC} & \textbf{SE} & \textbf{SP} & \textbf{ACC} \\ 
    \midrule
    U-Net~\cite{ronneberger2015u} & 0.8545 & 0.8800 & 0.9697 & 0.9404  
    \\
    Att-UNet~\cite{schlemper2019attention} & 0.8566 & 0.8674 & \textcolor{blue}{0.9863} & 0.9376 
    \\
    TransUNet~\cite{chen2021transunet} & 0.8499 & 0.8578 & 0.9653 & 0.9452  
    \\
    MCGU-Net~\cite{asadi2020multi}  & 0.8950 & 0.8480 & \textcolor{red}{0.9860}  & 0.9550 
    \\
    MedT~\cite{valanarasu2021medical} & 0.8389 & 0.8252 & 0.9637 & 0.9358  
    \\
    FAT-Net~\cite{wu2022fat}  & 0.8903 & \textcolor{red}{0.9100} & 0.9699 & 0.9578 
    \\
    TMU-Net~\cite{reza2022contextual}  & 0.9059 & 0.9038 & 0.9746 & 0.9603  
    \\
    Swin-Unet~\cite{cao2021swinunet}  & 0.8946 & 0.9056 & 0.9798 & 0.9645
    \\
    \midrule
    Baseline & 0.8863 & 0.8852 & 0.9592 & 0.9478 
    \\
    FET-W{$\star$} & \textcolor{red}{0.9085} & \textcolor{blue}{0.9123} & 0.9805 & \textcolor{red}{0.9601} 
    \\
    \rowcolor[HTML]{C8FFFD}
    \textbf{FET} & \textcolor{blue}{0.9157} & 0.8900 & 0.9827 & \textcolor{blue}{0.9690} 
    \\
    \bottomrule
    \end{tabular}
    }\\ \vspace{0.2em}
    {\tiny$\star$FET-W: FET (without MSCE bridge)}
    
    \end{subtable}
\end{table}

% \begin{figure}[!tbh]
%     \centering
%     % \includegraphics[width = \textwidth]{./Figures/synapsevisualization_wavelet.pdf}
%     \includegraphics[width = \textwidth]{./Figures/synapsevisualization_wavelet_2.pdf}
%     \caption{Segmentation results of the proposed method versus SOTA methods on the \textit{Synapse} dataset.}
% 	\label{fig:synapseviz}
% \end{figure}

\begin{figure}[!hb]
    \centering
    \includegraphics[width=0.81\textwidth]{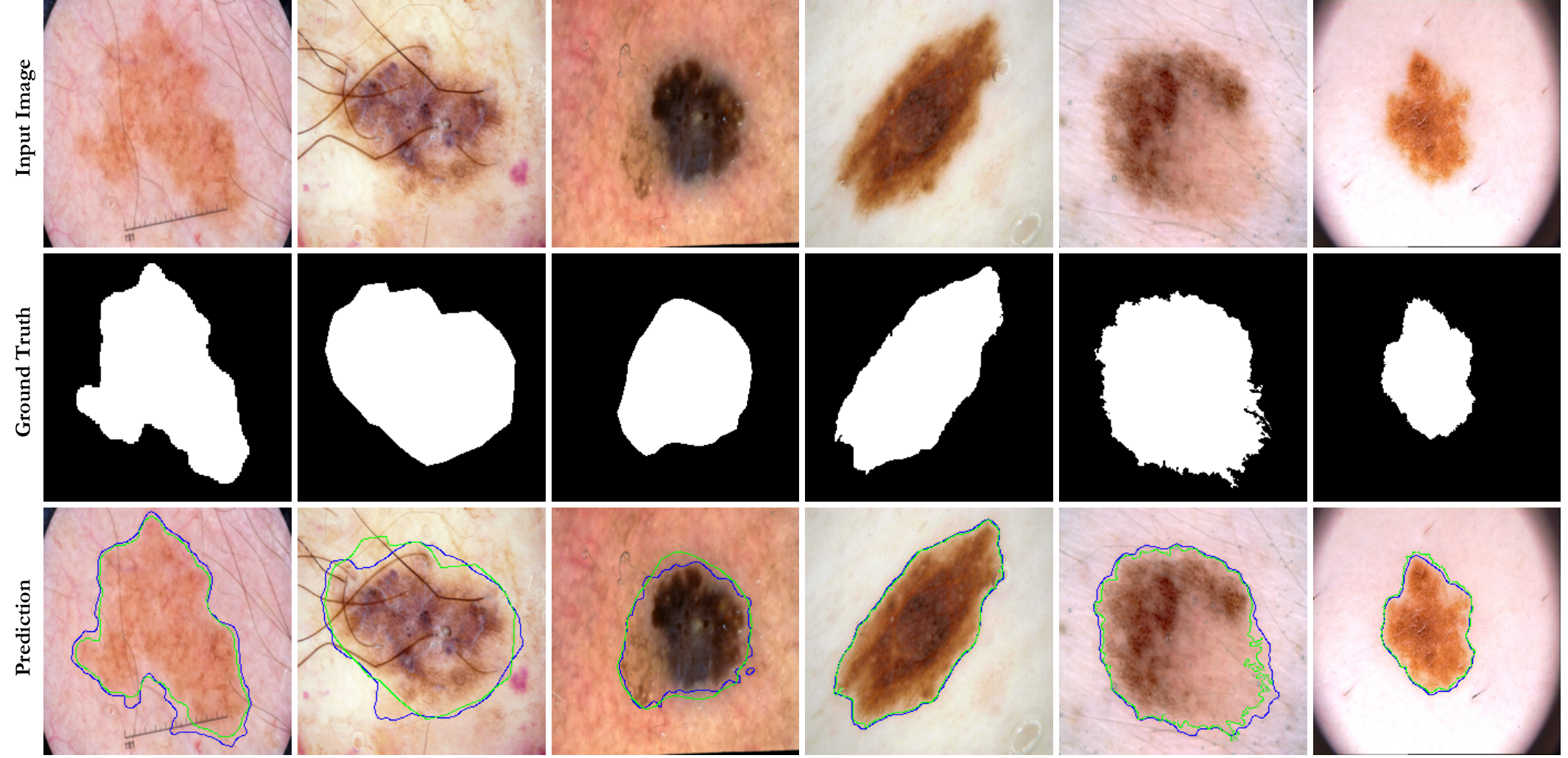}
    \caption{Visual representation of FET model on the \textit{ISIC 2018}dataset. Ground truth and prediction boundaries are shown in \textcolor{green}{green}, and \textcolor{blue}{blue} colors, respectively.}
    \label{fig:skin_visualization}
\end{figure}

\noindent\textbf{Skin Lesion Segmentation:}
\Cref{tab:skin_comparison} also endorses the capability of FET compared to other well-known methods for skin lesion segmentation methods. Specifically, our method performs better than hybrid approaches such as TMU-Net~\cite{reza2022contextual}. Additionally, our method proves to be more resilient to noisy elements when compared to pure Transformer-based methods such as Swin-Unet~\cite{cao2021swinunet}, which suffer from reduced performance due to a lack of emphasis on local texture modeling. In addition, comparing qualitative results (presented in \Cref{fig:skin_visualization}) on the ISIC 2018 dataset approves our method's capability to capture fine-grained boundary information.

\begin{figure}[!h]
    \centering
    {   
        \rotatebox{90}{\tiny \hspace{15pt} Standard}
        \rotatebox{90}{\tiny \hspace{15pt} Self-Att}
    }
    \begin{subfigure}[b]{0.3135\textwidth}
        \centering
        \includegraphics[width=\textwidth]{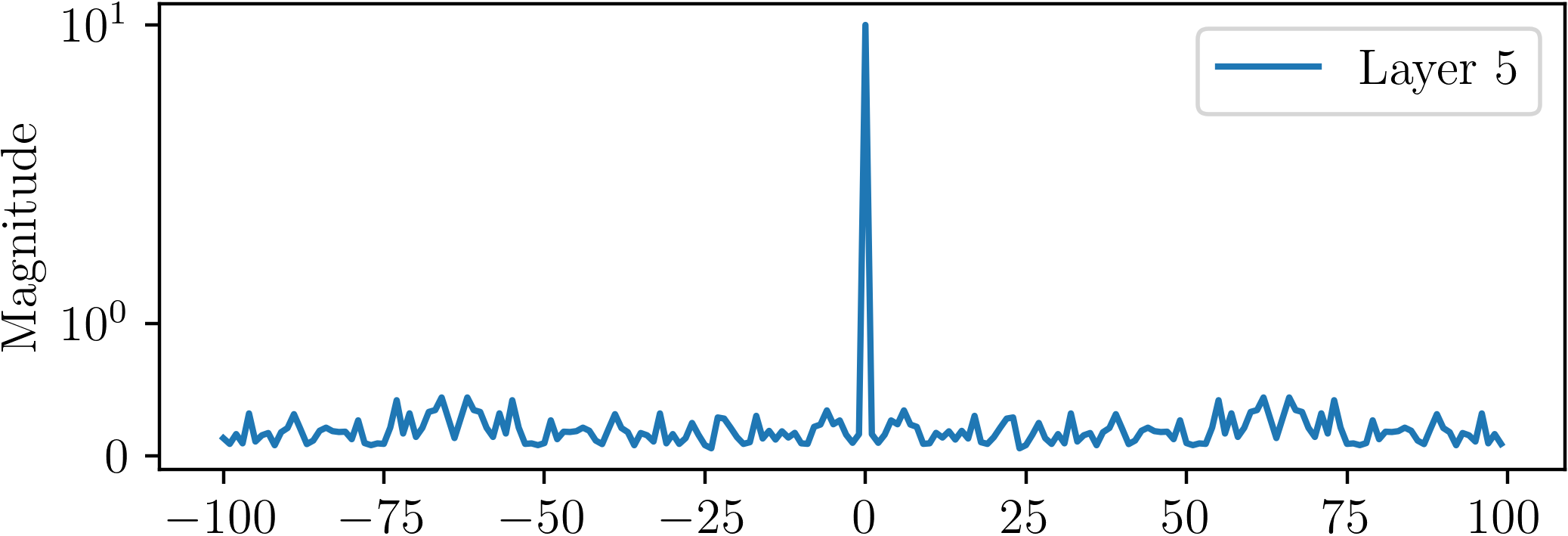}
        % \caption{spec s1}
        \label{fig:vis_spec_s1}
    \end{subfigure}
    \hfill
    \begin{subfigure}[b]{0.3135\textwidth}
        \centering
        \includegraphics[width=\textwidth]{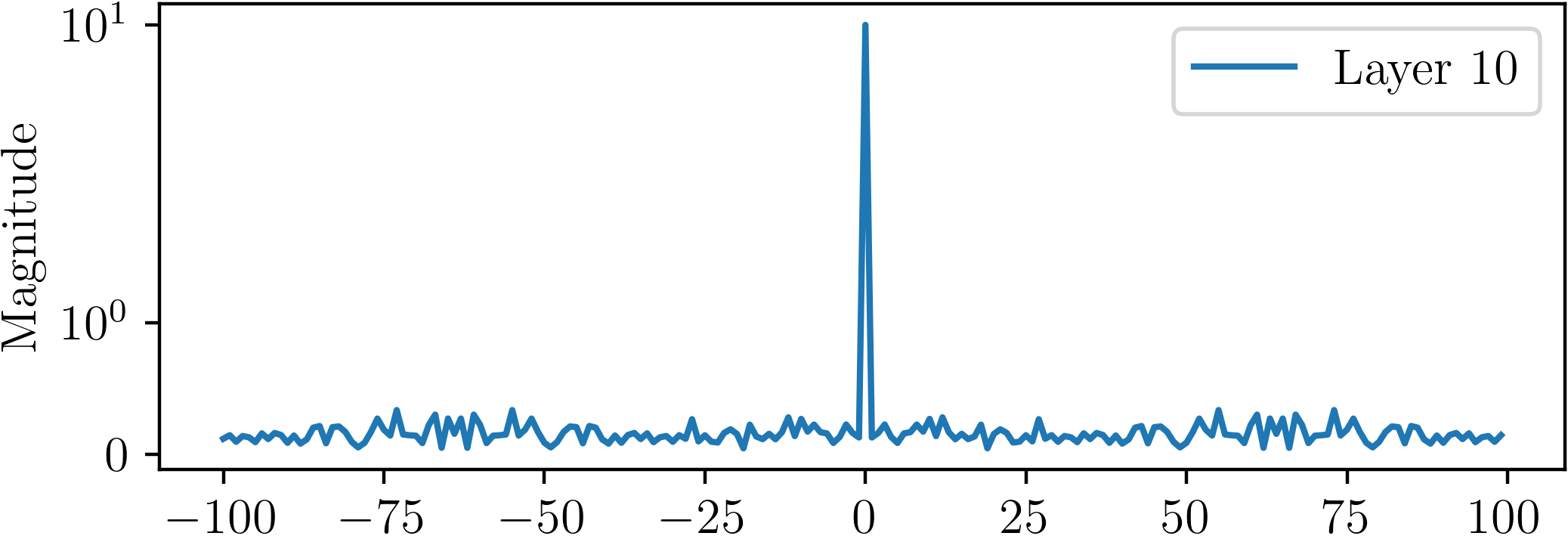}
        % \caption{spec s2}
        \label{fig:vis_spec_s2}
    \end{subfigure}
    \hfill
    \begin{subfigure}[b]{0.3135\textwidth}
        \centering
        \includegraphics[width=\textwidth]{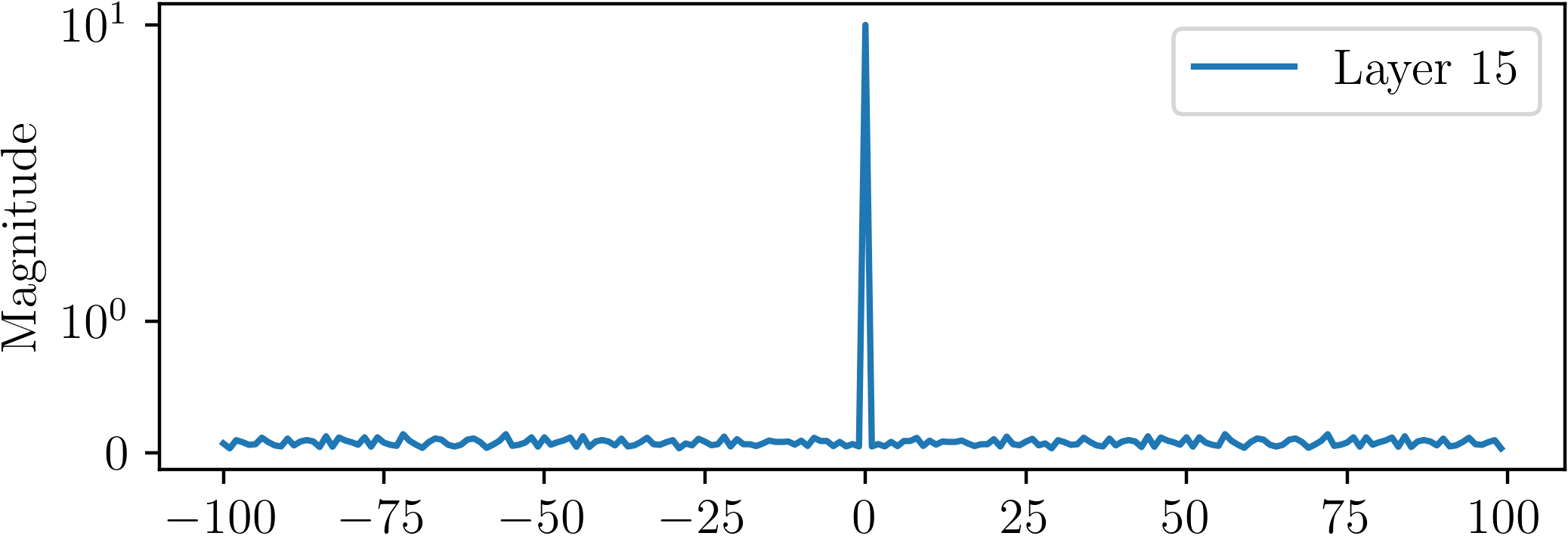}
        % \caption{spec s3}
        \label{fig:vis_spec_s3}
    \end{subfigure}
    \\ \vspace{-8pt}
    {
        \rotatebox{90}{\tiny \hspace{15pt} FET}
        \rotatebox{90}{\tiny \hspace{15pt} Self-Att}
    }
    \centering
    \begin{subfigure}[b]{0.3135\textwidth}
        \centering
        \includegraphics[width=\textwidth]{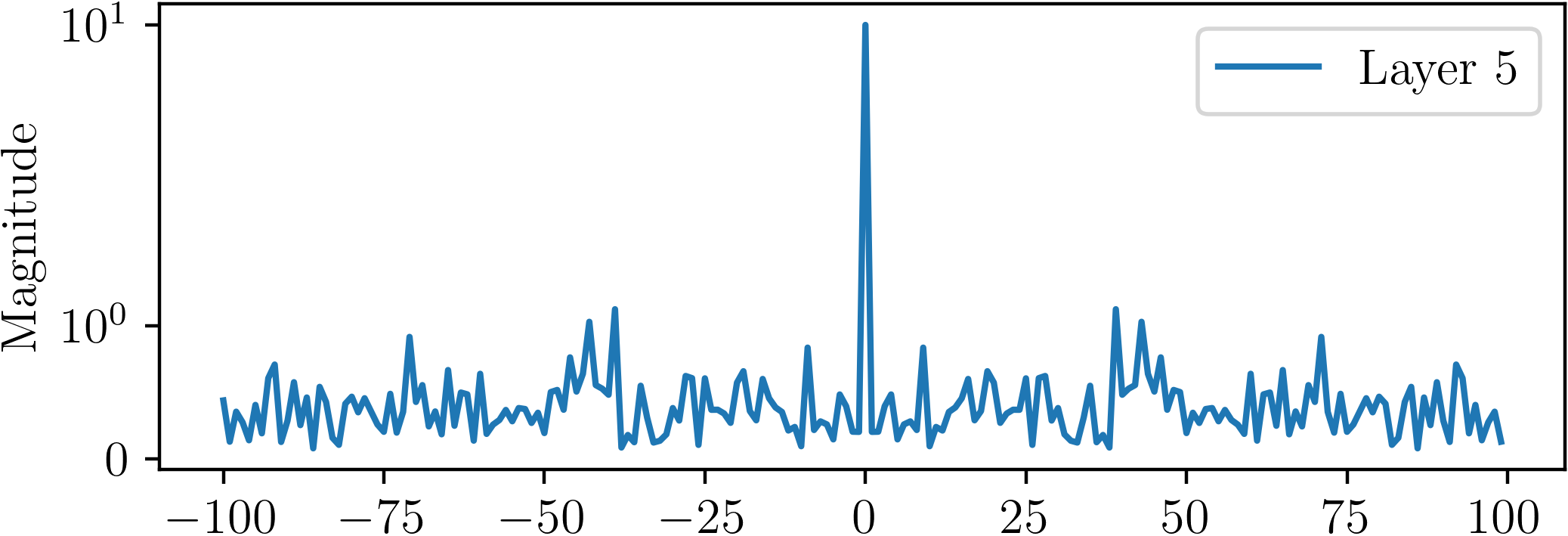}
        % \caption{spec w1}
        \label{fig:vis_spec_w1}
    \end{subfigure}
    \hfill
    \begin{subfigure}[b]{0.3135\textwidth}
        \centering
        \includegraphics[width=\textwidth]{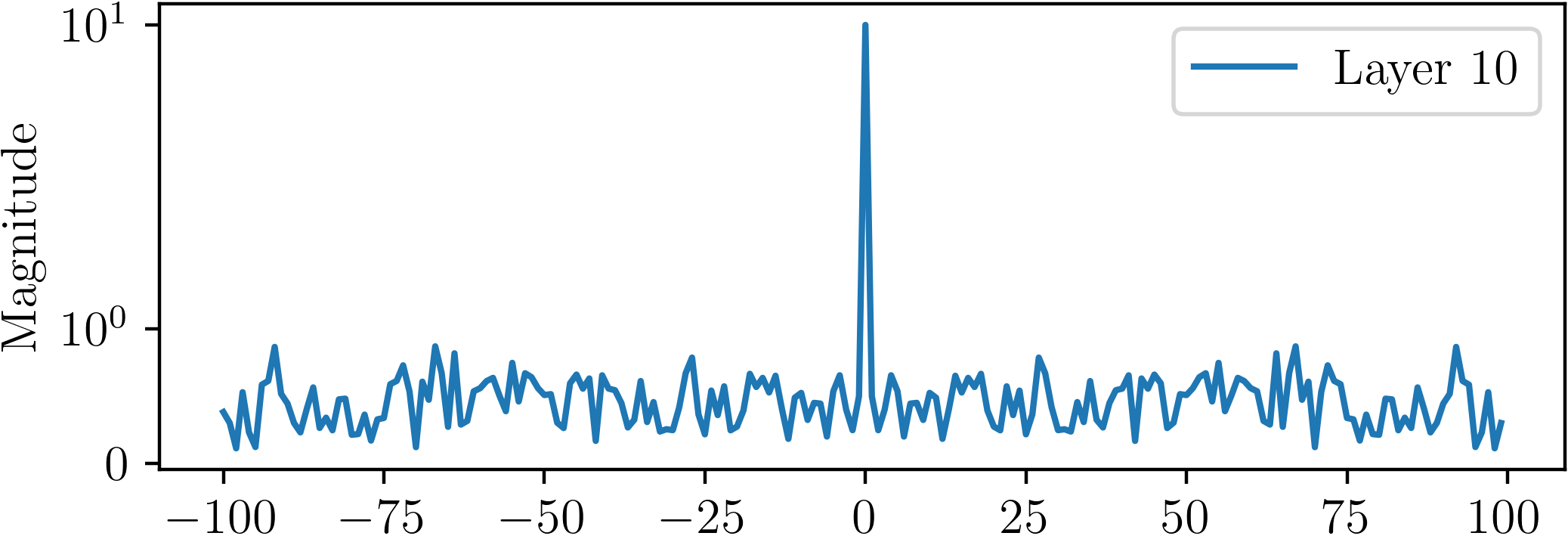}
        % \caption{spec w2}
        \label{fig:vis_spec_w2}
    \end{subfigure}
    \hfill
    \begin{subfigure}[b]{0.3135\textwidth}
        \centering
        \includegraphics[width=\textwidth]{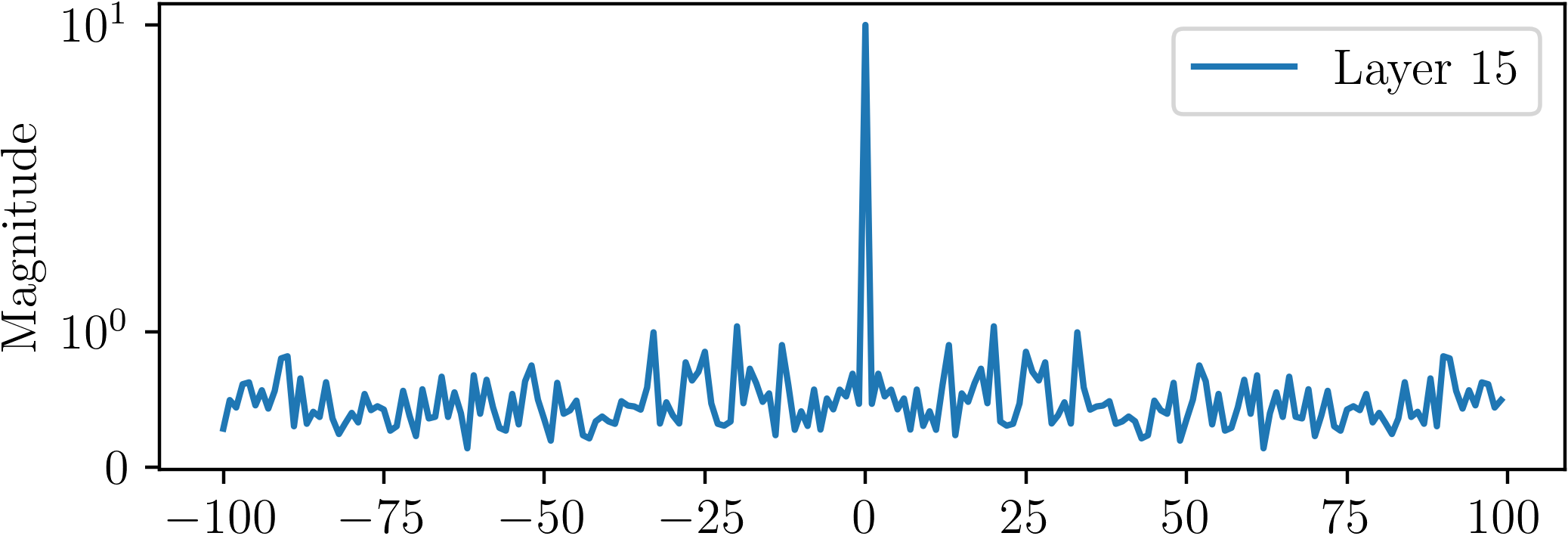}
        % \caption{spec w3}
        \label{fig:vis_spec_w3}
    \end{subfigure}
    \\ \vspace{-1.8em}
    \caption{Illustration of the spectral response of Standard Transformer (up) and FET (down) for capturing different frequency representation.}
    % The magnitude of the Discrete Fourier Transform (DFT) representation of output attention maps in log scale depicts that by increasing depth, the performance of the Standard Transformer degrades dramatically in the high-frequency capturing term. In contrast, FET performs seamlessly in capturing high frequency in different layers.}
    \label{fig:spectrum_visualization}
\end{figure}

To comprehensively evaluate the influence of our module on capturing high-frequency information in deeper layers, we conducted an extensive analysis of the spectrum response in \Cref{fig:spectrum_visualization}. Our findings reveal that our method stands out from traditional self-attention modules by effectively preserving high-frequency information within the depths of the network.

To further assess the effectiveness of our approach in capturing both local and global information, we have included the visualization of attention maps in \Cref{fig:spectrum_visualization}. The results clearly demonstrate our method's capability to successfully detect both small and large organs.

\begin{figure}[th]
\centering
\resizebox{\textwidth}{!}{
    \begin{tabular}{@{} *{6}c @{}}
    \includegraphics[width=0.25\textwidth]{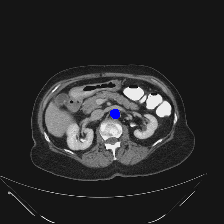} &
    \includegraphics[width=0.25\textwidth]{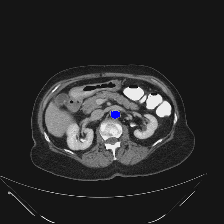} &
    \includegraphics[width=0.25\textwidth]{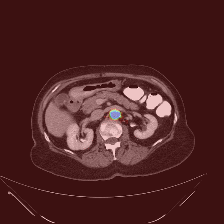} &
    \includegraphics[width=0.25\textwidth]{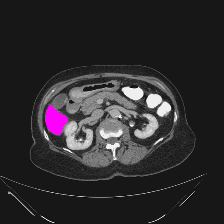} &
    \includegraphics[width=0.25\textwidth]{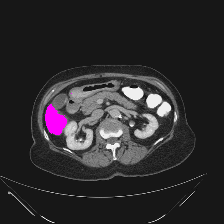} &
    \includegraphics[width=0.25\textwidth]{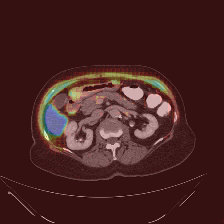} \\
    \includegraphics[width=0.25\textwidth]{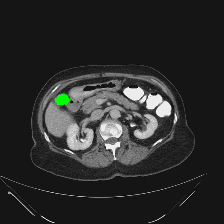} &
    \includegraphics[width=0.25\textwidth]{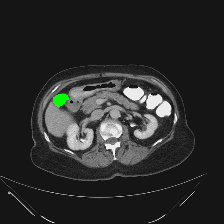} &
    \includegraphics[width=0.25\textwidth]{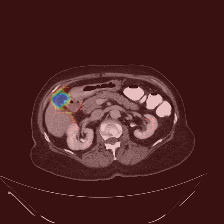} &
    \includegraphics[width=0.25\textwidth]{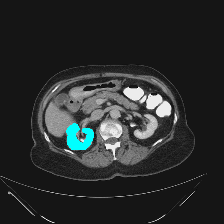} &
    \includegraphics[width=0.25\textwidth]{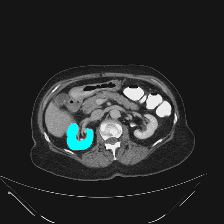} &
    \includegraphics[width=0.25\textwidth]{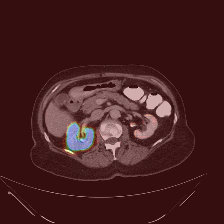} \\
    {\small (a) Ground Truth} & {\small(b) Prediction} & {\small(c) Heatmap} & {\small (d) Ground Truth} & {\small(e) Prediction} & {\small(f) Heatmap}
    \end{tabular}
}
\caption{The performance of the FET model was evaluated by visualizing its attention map using Grad-CAM on the \textit{Synapse} dataset. The results showed that the model effectively detects both small (\textit{i.e.}, aorta and gallbladder from the right side's top to bottom) and large organs (\textit{i.e.}, liver and right kidney from the left side's top to bottom), demonstrating its effectiveness in capturing long-range dependencies and local features. In summary, the FET model performed well in detecting organs on the \textit{Synapse} dataset.} \label{fig:heat_map}
\end{figure}

\section{Conclusion}\label{sec:conclusion}
In this paper, we redesigned the Transformer block to recalibrate spatial and context representation adaptively. We further imposed a secondary attention map to highlight the importance of boundary information within the Transformer block. Moreover, we modeled the intra-scale dependency for further performance improvement by redesigning the skip connection path. The effectiveness of our module is illustrated through the experimental results. \\

\textbf{Acknowledgments:} This work was funded by the German Research Foundation (Deutsche Forschungsgemeinschaft, DFG) under project number 191948804.
We would like to thanks Elnaz Khorami for her guidence on the proposed method and mathematical formulation.

\bibliographystyle{splncs04}
\bibliography{MLMI57.bib}
\end{document}